\documentclass[10pt,twocolumn,letterpaper]{article}

\usepackage[pagenumbers]{cvpr} %
\usepackage{times}
\usepackage{epsfig}
\usepackage{graphicx}
\usepackage{amsmath}
\usepackage{amssymb}
\usepackage{multirow}
\usepackage{booktabs}
\usepackage[dvipsnames]{xcolor}

\usepackage[pagebackref,breaklinks,colorlinks]{hyperref}

\usepackage[capitalize]{cleveref}
\crefname{section}{Sec.}{Secs.}
\Crefname{section}{Section}{Sections}
\Crefname{table}{Table}{Tables}
\crefname{table}{Tab.}{Tabs.}

\def\cvprPaperID{6051} %
\def\confName{CVPR}
\def\confYear{2022}
\def\name{ESeg}

\begin{document}

\title{Revisiting Multi-Scale Feature Fusion for Semantic Segmentation}

\author{
Tianjian Meng $^1\dagger$\qquad
Golnaz Ghiasi $^1$\qquad
Reza Mahjourian $^2$\qquad
Quoc V. Le $^1$\qquad
Mingxing Tan $^1\dagger$ \\
$^1$ Google Research\qquad
$^2$ Waymo Inc. \qquad \\
$\dagger$\{mengtianjian, tanmingxing\}@google.com
}

\maketitle

\begin{abstract}
   It is commonly believed that high internal resolution combined with expensive operations (e.g. atrous convolutions) are necessary for accurate semantic segmentation, resulting in slow speed and large memory usage. In this paper, we question this belief and demonstrate that neither high internal resolution nor atrous convolutions are necessary. Our intuition is that although segmentation is a dense per-pixel prediction task, the semantics of each pixel often depend on both nearby neighbors and far-away context; therefore, a more powerful multi-scale feature fusion network plays a critical role. Following this intuition, we revisit the conventional multi-scale feature space (typically capped at $P_5$\footnote{$P_i$ denotes features with resolution $P_0/2^i$, where $P_0$ is the input image size.}) and extend it to a much richer space, up to $P_9$, where the smallest features are only $1/512$ of the input size and thus have very large receptive fields. To process such a rich feature space, we leverage the recent BiFPN to fuse the multi-scale features. Based on these insights, we develop a simplified segmentation model, named \textbf{\name}, which has neither high internal resolution nor expensive atrous convolutions. Perhaps surprisingly, our simple method can achieve better accuracy with faster speed than prior art across multiple datasets. In real-time settings, {\name}-Lite-S achieves 76.0\% mIoU on CityScapes~\cite{Cordts2016Cityscapes} at 189 FPS, outperforming FasterSeg~\cite{chen2019fasterseg} (73.1\% mIoU at 170 FPS). Our {\name}-Lite-L runs at 79 FPS and achieves 80.1\% mIoU, largely closing the gap between real-time and high-performance segmentation models.
   
\end{abstract}

\begin{figure}[t]
	   \includegraphics[width=1.0\columnwidth]{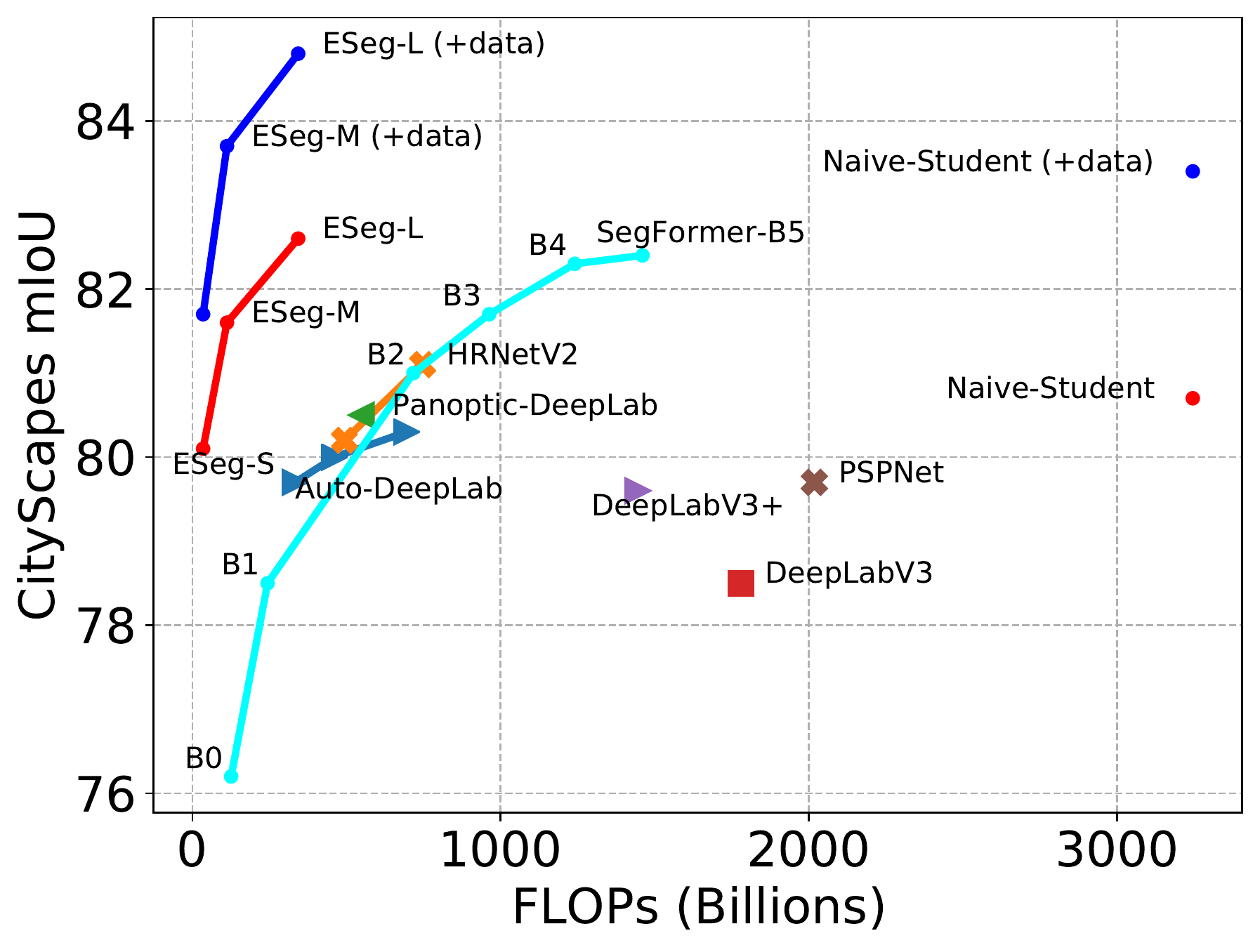}
	   	\vskip -0.1in
	   \caption{\textbf{Model Sizes vs. CityScapes validation mIoU}. All models in the figure are using single-scale evaluation protocol. \texttt{+data} denotes using extra data for pretraining and self-training. The FLOPs are calculated at 1024$\times$2048 input resolution. Our proposed {\name} models are much simpler, yet still outperform previous models by better quality and less computation cost.}
	   \vspace{-1em}
	\label{fig:teaser}
	\end{figure}

 \section{Introduction}
     
Semantic segmentation is an important computer vision task that has been widely used in robotics and autonomous driving. The main challenge in semantic segmentation lies in the per-pixel dense prediction, where we need to predict the semantic class for each pixel. State-of-the-art (SOTA) segmentation models~\cite{yu2017dilated,zhao2017pyramid,chen2017rethinking,chen2018encoder,sun2019high} heavily rely on high internal resolution for dense prediction, and it is commonly believed that such high internal resolution is necessary to learn accurate per-pixel semantics, leading to slow runtime speed and large runtime memory usage. For example, DeepLabV3+~\cite{chen2018encoder} requires more than 1000B FLOPs (vs common real-time models using $<$100B FLOPs), making it prohibitively expensive to apply to real-time scenarios. Meanwhile, the high-resolution feature maps make it difficult to obtain large receptive fields. Many attempts have been made to address this issue, for example by replacing the regular convolutions with atrous convolutions~\cite{yu2017dilated,zhao2017pyramid,chen2017rethinking,chen2018encoder} or adding additional attention modules~\cite{yuan2018ocnet,yuan2019object,fu2019adaptive} to enlarge receptive fields.
Although these methods improve accuracy, atrous convolutions and attention modules are usually much slower than regular convolutions.
     
Recent real-time semantic segmentation models tend to learn high-resolution representation in different ways to satisfy the latency constraints, either by hand-crafting better networks~\cite{yu2020bisenet,hu2020real} or by using neural architecture search~\cite{chen2019fasterseg,li2019partial}. These methods tend to improve accuracy, but they still need to maintain the high internal resolution.

In this paper, we question the belief in high internal resolution and demonstrate that neither high internal resolution nor atrous convolutions are necessary for accurate segmentation. Our intuition is that although segmentation is a dense per-pixel prediction task, the semantics of each pixel often depend on both nearby neighbors and far-away context; therefore, conventional multi-scale feature space can easily be a bottleneck, and a more powerful multi-resolution feature fusion network is desired. Following this intuition, we revisit the conventional simple feature fusion space (typically up to $P_5$ from traditional classification backbones) and extend it to a much richer space, up to $P_9$, where the smallest internal resolution is only $1/512$ of the input image size and thus can have a very large receptive field. 

Based on the richer multi-resolution feature space, we present a simplified segmentation model named {\name}, which has neither high internal resolution nor expensive atrous convolutions. To keep the model as simple as possible, we adopt the naive encoder-decoder network structure, where the encoder is an off-the-shelf EfficientNet~\cite{tan2019efficientnet} and the decoder is a slightly-modified BiFPN~\cite{tan2020efficientdet}. We observe that as the multi-resolution feature space up to $P_9$ has much richer information than conventional approaches, a powerful bi-directional feature pyramid is extremely important here to effectively fuse these features. Unlike the conventional top-down FPN, BiFPN allows both top-down and bottom-up feature fusion, enabling more effective feature fusion in the rich multi-scale feature space.

We evaluate our {\name} on CityScapes~\cite{Cordts2016Cityscapes} and ADE20K~\cite{zhou2017scene}, two widely used benchmarks for semantic segmentation. Surprisingly, although {\name} is much simpler, it consistently matches the performance of prior art across different datasets. By scaling up the network size, our {\name} models achieve better accuracy while using 2$\times$ - 9$\times$ fewer parameters and 4$\times$ - 50$\times$ fewer FLOPs than competitive methods. In real-time settings, our {\name}-Lite-S achieves 76.0\% mIoU on the CityScapes validation set at 189 FPS, outperforming the prior art of FasterSeg by 2.9\% mIoU at faster speed. With 80.1\% CityScapes mIoU at 79 FPS, our {\name}-Lite-L bridges the gap between real-time and advanced segmentation models for the first time.

In addition to the standard CityScapes/ADE20K datasets, our models also perform well on large-scale datasets. By pretraining on the larger Mapillaly Vistas~\cite{neuhold2017mapillary} and self-training on CityScapes coarse-labeled data, our {\name}-L achieves 84.8\% mIoU on the CityScapes validation set, while being much smaller and faster than prior art. These results highlight the importance of multi-scale features for semantic segmentation, demonstrating that a simple network is also able to achieve state-of-the-art performance. %

\begin{figure*}
\centering
	\includegraphics[width=0.85\linewidth]{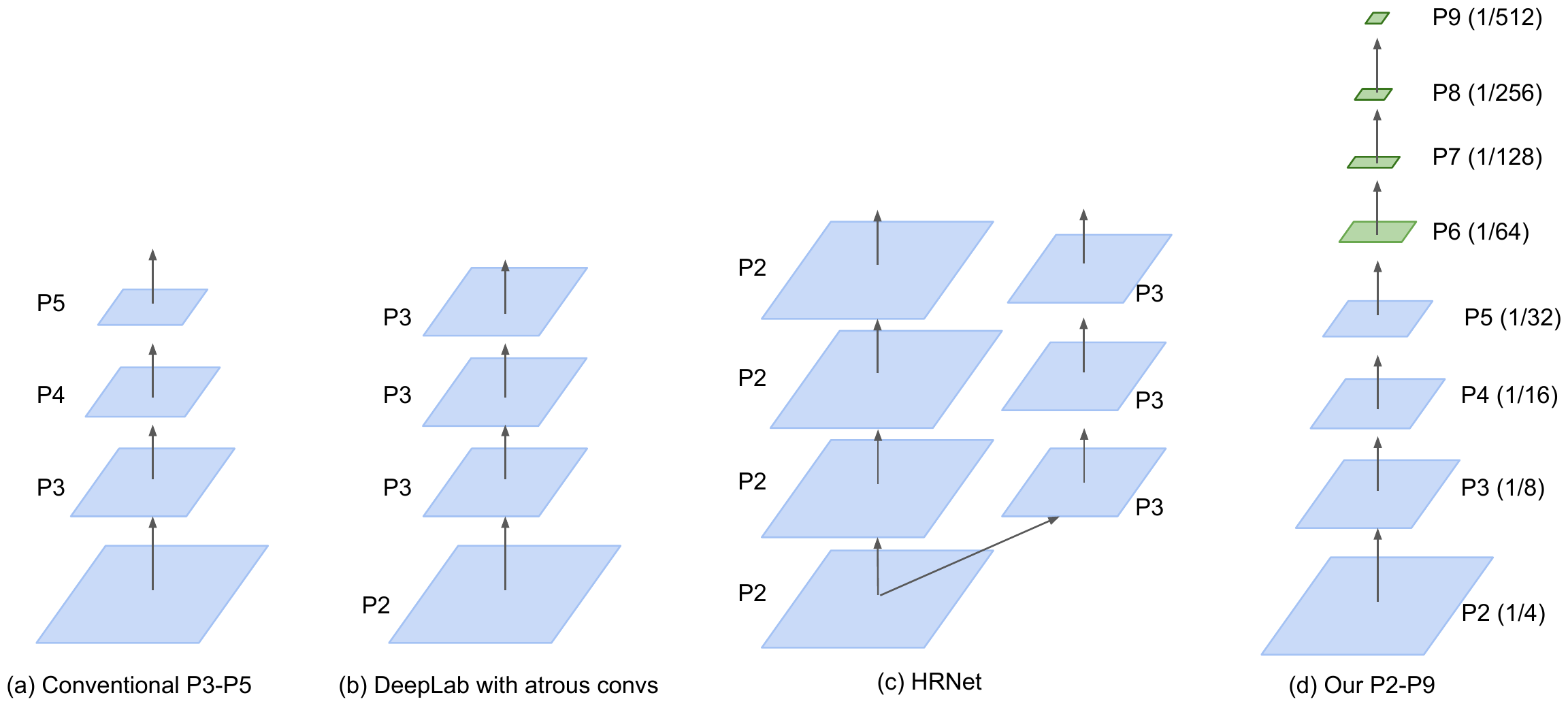}
	\vskip -0.1in
	\caption{\textbf{Multi-Scale Feature Space}. $P_i$ denotes a feature map with resolution $\dfrac{P_0}{2^i}$, where $P_0$ is the input image size. (a) conventional feature networks mostly use up to $P_5$ from an ImageNet backbone, but the accuracy is relatively low; (b) DeepLab\cite{chen2017deeplab,chen2018encoder}  maintains the high resolution and use atrous convolutions to capture larger receptive fields, with the overhead of slow runtime; (c) HRNet~\cite{sun2019high} maintains a separate path for each resolution, resulting in large memory usage for high-resolution features; (d) we simply add more low-resolution features $P_6-P_9$ to capture high-level semantics. The additional $P_6-P_9$ features improves accuracy (see Table \ref{tab:level}) with very minimal overhead, thanks to large receptive fields on the low-resolution feature maps.
	}
\vspace{-1em}
\label{fig:featurespace}
\end{figure*}

\section{Revisiting Multi-Scale Feature Fusion}
Semantic segmentation models need to predict the semantic class for individual pixels. Due to this dense prediction requirement, it is common to maintain high-resolution features and perform pixel-wise classification on them. However, high-resolution features bring two critical challenges: First, they often require expensive computation and large memory, at $O(n^2)$ complexity with respect to the resolution; Second, it is difficult to obtain large receptive fields with regular convolutions, so capturing long-distance correlations would require very deep networks, expensive atrous operations, or spatial attention mechanisms, making them even slower and difficult to train.

To mitigate the issues of high-resolution features, many previous works employ multi-scale feature fusion. The core idea is to leverage multiple features with different resolutions to capture both short- and long-distance patterns. Specifically, a common practice is to adopt an ImageNet-pretrained backbone network (such as ResNet~\cite{he2016deep}) to extract feature scales up to $P_5$, and then apply an FPN~\cite{lin2017feature} to fuse these features in a top-down manner, as illustrated in Figure \ref{fig:featurespace}(a). However, recent works~\cite{chen2017deeplab,sun2019high} show such naive feature fusion is inferior to high resolution features: For example, DeepLab tends to keep high-resolution features combined with atrous convolutions as shown in Figure~\ref{fig:featurespace}(b), but atrous convolutions are usually hardware-unfriendly and slow. Similarly, HRNet~\cite{sun2019high} maintains multiple parallel branches of high-resolution features as shown in Figure~\ref{fig:featurespace}(c). Both approaches lead to high computational cost and memory usage.

\begin{table}[!h]
\begin{center}
\resizebox{0.85\columnwidth}{!}{
\begin{tabular}{c|lll}
\hline
Feature Levels & mIoU & \#Params & FLOPs \\
\hline
$P_2-P_5$ & 78.3 & 6.4M & 34.3B \\
\hline
$P_2-P_7$ & 79.5 (\textcolor{blue}{+1.2}) & 6.6M & 34.5B \\
\hline
$P_2-P_9$ & 80.1 (\textcolor{blue}{+1.8}) & 6.9M & 34.5B \\
\hline
\end{tabular}
}
\end{center}
\vskip -0.1in
\caption{\textbf{Comparison of different feature levels on CityScapes dataset}. $P_i$ feature level has resolution of $P_0*2^{-i}$, where $P_0$ is the input image size. Use more high-level features would enlarge receptive fields and improve accuracy with negligible overhead.}
\label{tab:level}
\end{table}

Here we revisit the multi-scale feature map and identify that insufficient feature levels are the key limitation of the vanilla $P_2-P_5$ multi-scale feature space. $P_2-P_5$ is originally designed for ImageNet backbones, where the image size is as small as 224$\times$224, but segmentation images have much higher resolution: in CityScapes, each image has a resolution of 1024$\times$2048. Such large input images require much larger receptive fields, raising the need for higher-level feature maps.  Table~\ref{tab:level} compares the performance of different feature levels. When we simply increase the feature levels to $P_2-P_9$ (adding four more extra level $P_6-P_9$), it significantly improves the performance by 1.8 mIoU with negligible overhead on parameters and FLOPs. Notably, for high level features ($P_6-P_9$) that are already down-sampled from the input image, there is no need to use hardware-unfriendly atrous convolutions to increase the receptive fields any more. This leads to our first observation:

\paragraph{Insight 1:} \emph{Larger feature spaces up to $P_9$ are much more effective than conventional spaces capped at $P_5$.}

Given such large feature space like $P_2-P_9$, a natural question is how to fuse them effectively. Many previous works simply apply a top-down FPN~\cite{lin2017feature} to fuse the features. Recently, PANet~\cite{liu2018path} proposes to add an extra bottom-up fusion path, and EfficientDet~\cite{tan2020efficientdet} proposes to fuse features in bidirectional fashion for object detection. Here we study the performance of different feature fusion approaches. As shown in Table~\ref{tab:bifpn}, when the feature space is larger ($P_2-P_9$), a more powerful BiFPN performs significantly better than a simple FPN. Intuitively, a larger feature space needs more layers to propagate the information and more bidirectional flows to effectively fuse them.  Based on these studies, we have the second observation:

\paragraph{Insight 2:} \emph{A powerful feature fusion network is critical for a large multi-scale feature space.}

\begin{table}
\begin{center}
\resizebox{0.99\columnwidth}{!}{
\begin{tabular}{cc|lll}
\hline
Feature Space & Feature Network & mIoU & FLOPs\\
\hline
\multirow{2}{*}{$P_2-P_5$} & FPN & 78.0 & 39.8B \\
& BiFPN & 78.3 (\textcolor{blue}{+0.3}) & 34.3B \\
\hline
\multirow{2}{*}{$P_2-P_9$} & FPN & 79.1 & 40.0B \\
& BiFPN & 80.1 (\textcolor{blue}{+1.0}) & 34.5B \\
\hline
\end{tabular}
}
\end{center}
\vskip -0.1in
\caption{\textbf{Comparison of different feature fusion networks for CityScapes dataset.} BiFPN and FPN has similar performance for small feature space $P_2-P_5$, but the different is much significant for large feature space $P_2-P_9$. We adjust the filter sizes of FPN and BiFPN to ensure they have comparable computational cost.
}
\label{tab:bifpn}
\vspace{-1em}
\end{table}
\section{{\name}}

\begin{figure*}
\centering
	\includegraphics[width=0.75\linewidth]{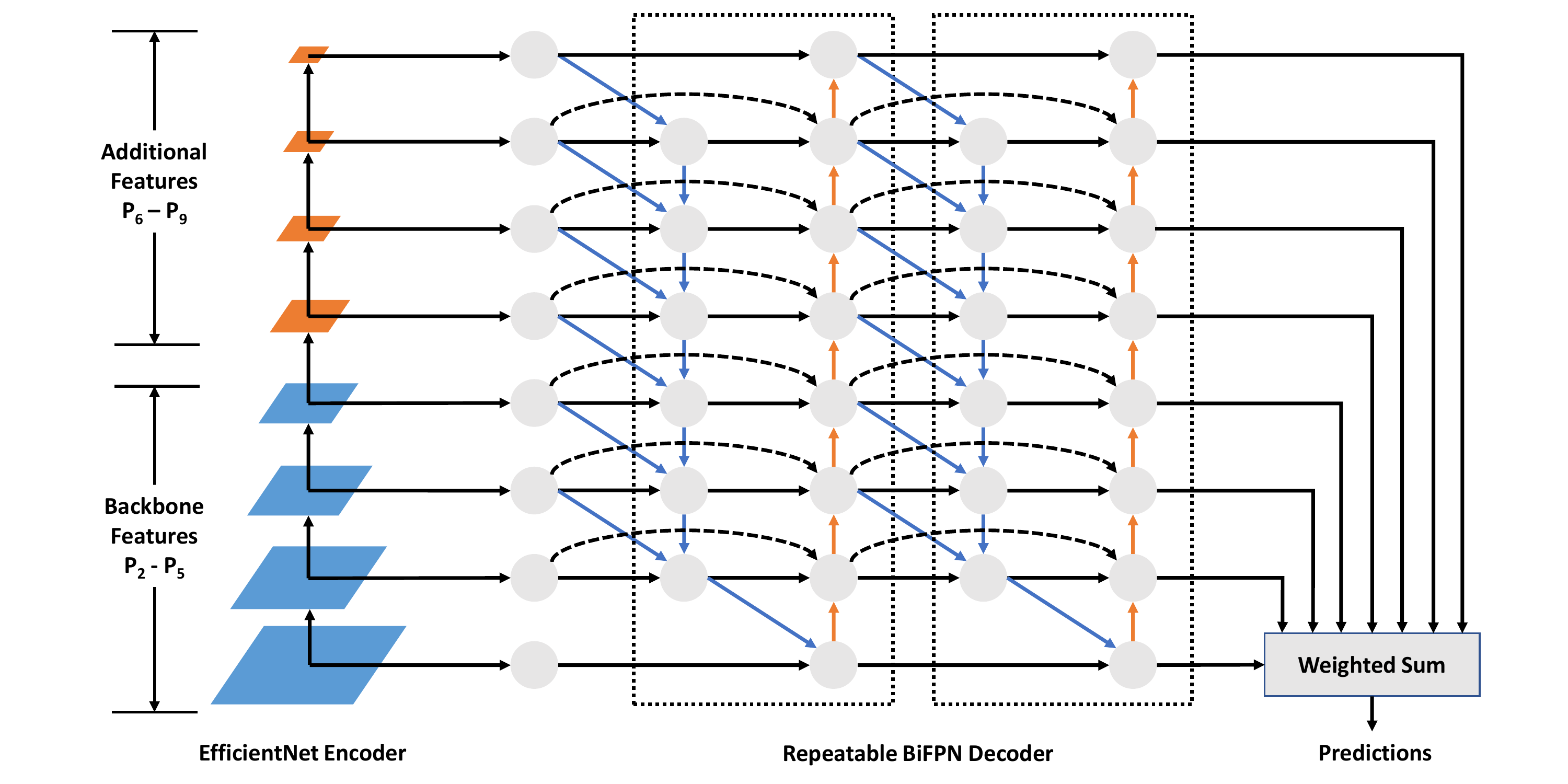}
\vskip -0.1in
\caption{\textbf{{\name} network architecture}. The  backbone~\cite{tan2019efficientnet} extracts \{$P_2 - P_5$\} feature maps from the raw input images; Four additional feature maps \{$P_6 - P_9$\} are added on top of these backbone features with simple average pooling. The decoder perform bidirectional multi-scale feature fusion~\cite{tan2020efficientdet} to strength the internal representations for each feature map. All feature maps are upsampled and combined with weighted sum to generate the final per-pixel prediction.}
\vspace{-1em}
\label{fig:arch}
\end{figure*}

Based on our two key insights, we now develop a simple encoder-decoder-based segmentation model, aiming to achieve better performance without relying on atrous convolutions or dedicated high-resolution features.

\subsection{Network Design}

Figure~\ref{fig:arch} shows an overview of our {\name} network. It has a standard encoder-decoder structure, where the encoder extracts multiple levels of features from the raw images and the decoder performs extensive multi-scale feature fusion. After the decoder, we upsample and weighted-sum all features, and predict the dense per-pixel semantic class. We describe each component and highlight our design choices in following:

\paragraph{Encoder: } \label{sec:encoder}
We use EfficientNet~\cite{tan2019efficientnet} as the backbone to extract the first five levels of features \{$P_1$, $P_2$, $P_3$, $P_4$, $P_5$\}, where $P_i$ denotes the $i$-th level of features with spatial size of $P_0 / 2^{i}$. Please note that our design is general and can be applied to any potentially-better convolutional classification backbones in the future. Here we use EfficientNet as our default backbone to illustrate our idea. Based on our \emph{insight 1}, we add four more extra features $P_6-P_9$ on top of $P_5$, by downsampling the features with 3$\times$3 stride-2 pooling layers. For simplicity, we use the same hidden size of $P_5$ for all other $P_6-P_9$ features. Notably, we only use regular convolutional and pooling layers in the encoder, without any use of hardware-unfriendly atrous convolutions.

\paragraph{Decoder: }
The decoder is responsible to combining the multi-scale features. Based on our \emph{insight 2}, we employ the powerful BiFPN~\cite{tan2020efficientdet} to repeatedly apply top-down and bottom-up bidirectional feature fusion. The BiFPN outputs the same shape of $P_2-P_9$ feature maps, but each of them is already fused with information from other feature scales.

\paragraph{Prediction:}
To generate the final prediction, we upsample all decoder features to a fixed resolution, and do a weighted sum to combine them into a unified high-resolution feature map. Specifically, given $P_2 - P_9$ feature maps, we introduce eight additional learnable weights $w_2 - w_9$, where $w_i$ is a scalar variable for input feature $P_i$. These features are then combined using a softmax-based weighted sum fusion: %

$$O = \sum_{i} \cfrac{e^{w_i}}{\sum_{j} e^{w_j}} \cdot \text{Upsample}(P_i)$$

\noindent where $w_i$ is a non-negative variable initialized as 1.0 and updated during back-propagation, and upsample is a simple bilinear interpolation operation to match the resolution. Intuitively, $w_i$ represents the importance of feature map $P_i$: if $P_i$ is more important to the output, then $w_i$ will become larger during training. After training is done, we can get rid of the softmax function and perform normal weighted-sum fusion for faster inference.

A prediction head is applied to the final output feature $O$ to produce the final pixel-level class predictions. Notably, since the raw input image size is usually very large (e.g., 1024$\times$2048), it is expensive to directly perform prediction at the resolution of input images. Instead, we apply our prediction head at the minimum feature level $P_2$ (one fourth of the original size), and then bilinearly upsample the prediction to the original input image resolution to calculate the per-pixel loss.

\subsection{Network and Data Scaling}

Our goal is to design a family of segmentation models with different accuracy and latency trade-offs; However, hand-crafting new models for each latency constraint is difficult and expensive. Here we take a different approach similar to previous compound scaling~\cite{tan2019efficientnet,tan2020efficientdet}.

For the encoder backbone, we use the same scaling factors as the original EfficientNet such that we can reuse its ImageNet checkpoints. For the the BiFPN decoder, we scale them up linearly by increasing the depth (number of repeats) and width (number of input/output channels).  Table~\ref{tab:scaling} shows the width and depth scaling configurations, where the smallest model {\name}-Lite-S has 17B FLOPs and the largest model {\name}-L has 342B FLOPs. By scaling up {\name}, we show that it can consistently outperform prior works across different network size constraints.

\begin{table}[!h]
\begin{center}
\resizebox{0.95\columnwidth}{!}{
\begin{tabular}{l|cccc}
\hline
\multirow{2}{*}{Model} & \multicolumn{2}{c}{Encoder} & \multicolumn{2}{c}{Decoder} \\
& Width & Depth & \# channels & \# repeats \\
\hline
{\name}-Lite-S & 0.4 & 0.6 & 64 & 1 \\
\hline
{\name}-Lite-M & 0.6 & 1.0 & 80 & 2 \\
\hline
{\name}-Lite-L & 1.0 & 1.0 & 96 & 3 \\
\hline
{\name}-S & 1.0 & 1.1 & 96 & 4 \\
\hline
{\name}-M & 1.4 & 1.8 & 192 & 5 \\
\hline
{\name}-L & 2.0 & 3.1 & 288 & 6 \\
\hline
\end{tabular}
}
\end{center}
\vskip -0.2in
\caption{\textbf{Network size scaling configurations}.}
\label{tab:scaling}
\vspace{-1em}
\end{table}

Since segmentation datasets are usually small, another scaling dimension is data size. In our experiments, we will show that by pretraining and self-training on extra data, we are also able to further improve model accuracy. We explicitly note if a model is trained with extra data.

\subsection{GPU Inference Optimization}

As many segmentation models run on GPU with TensorRT, we further optimize our real-time {\name}-Lite models when running with TensorRT. First, we replace all MBConv blocks~\cite{sandler2018mobilenetv2,tan2019efficientnet} with Fused-MBConv~\cite{gupta2019efficientnet}, as depthwise convolutions are not well supported in TensorRT and cannot fully utilize GPU parallelisms. Second, we also remove GPU-unfriendly operations such as squeeze-and-excitation~\cite{hu2018squeeze}, and replace the SiLU(Swish-1)~\cite{elfwing2018sigmoid,ramachandran2017searching} activation with regular ReLU. Lastly, we further increase the base level features from $P_2$ to $P_3$ to avoid the expensive computations on large spatial dimensions, and move the prediction head to $P_3$ accordingly.  %
\begin{table*}
\begin{center}
\begin{tabular}{l|cccccc}
\hline
\multirow{2}{*}{Model} & val mIoU & val mIoU & \multirow{2}{*}{Params} & \multirow{2}{*}{Ratio} & \multirow{2}{*}{FLOPs} & \multirow{2}{*}{Ratio} \\
& w/o extra data & w/ extra data & & & & \\
\hline
\textbf{{\name}-S} & \textbf{80.1} & \textbf{81.7} & \textbf{6.9M} & \textbf{1x} & \textbf{34.5B} & \textbf{1x} \\
Auto-DeepLab-S~\cite{liu2019auto} & 79.7 & - & 10.2M & 1.5x & 333B & 9.7x \\
PSPNet (ResNet-101)~\cite{zhao2017pyramid} & 79.7 & - & 65.9M & 9.6x & 2018B & 59x \\
OCR (ResNet-101)~\cite{yuan2019object} & 79.6 & - & - & - & - & - \\
DeepLabV3+ (Xception-71)~\cite{chen2018encoder} & 79.6 & - & 43.5M & 6.3x & 1445B & 42x \\
DeepLabV3+ (ResNeXt-50)~\cite{zhu2019improving} & 79.5 & 81.4 & - & - & - & - \\
DeepLabV3 (ResNet-101)~\cite{chen2017rethinking} & 78.5 & - & 58.0M & 8.4x & 1779B & 52x \\
\hline
\textbf{{\name}-M} & \textbf{81.6} & \textbf{83.7} & \textbf{20.0M} & \textbf{1x} & \textbf{112B} & \textbf{1x} \\
HRNetV2-W48~\cite{sun2019high} & 81.1 & - & 65.9M & 3.3x & 747B & 6.7x \\
OCR (HRNet-W48)~\cite{yuan2019object} & 81.1 & - & - & - & - & - \\
ACNet (ResNet-101)~\cite{fu2019adaptive} & 80.9 & - & - & - & - & - \\
Naive-Student~\cite{chen2020semi} & 80.7 & 83.4 & 147.3M & 7.3x & 3246B & 29x \\
Panoptic-DeepLab (X-71)~\cite{cheng2020panoptic} & 80.5 & 82.5 & 46.7M & 2.3x & 548B & 4.9x \\
DeepLabV3 (ResNeSt-101)~\cite{zhang2020resnest} & 80.4$\dagger$ & - & - & - & - & - \\
Auto-DeepLab-L~\cite{liu2019auto} & 80.3 & - & 44.4M & 2.2x & 695B & 6.2x \\
HRNetV2-W40~\cite{sun2019high} & 80.2 & - & 45.2M & 2.3x & 493B & 4.1x \\
Auto-DeepLab-M~\cite{liu2019auto} & 80.0 & - & 21.6M & 1.1x & 461B & 4.1x \\
DeepLabV3 (ResNeSt-50)~\cite{zhang2020resnest} & 79.9$\dagger$ & - & - & - & - & - \\
OCNet (ResNet-101)~\cite{yuan2018ocnet} & 79.6 & - & - & - & - & - \\
\hline
\textbf{{\name}-L} & \textbf{82.6} & \textbf{84.8} & \textbf{70.5M} & \textbf{1x} & \textbf{343B} & \textbf{1x} \\
SegFormer-B5~\cite{xie2021segformer} & 82.4 & - & 84.7M & 1.2x & 1460B & 4.3x \\
\hline
\end{tabular}
\end{center}
\vskip -0.2in
\caption{\textbf{Performance comparison on CityScapes.} $\dagger$ denotes results using multi-scale evaluation protocol. All our models are evaluated in single-scale evaluation protocol.}
\label{tab:main}
\vspace{-1em}
\end{table*}

\section{Experiments}

We evaluate {\name} on the popular CityScapes~\cite{Cordts2016Cityscapes} and ADE20K~\cite{zhou2017scene} datasets, and compare the results with previous state-of-the-art segmentation models.

\subsection{Setup}

Our models are trained on 8 TPU cores with a batch size of 16. We use SGD as our optimizer with momentum 0.9.
Following previous work~\cite{sun2019high}, we apply random horizontal flipping and random scale jittering [0.5, 2.0] during our training. We use cosine learning rate decay, which yields similar performance as previous polynomial learning rate decay but with less tunable hyperparameters.
For fair comparison with \cite{yuan2019object,sun2019high,cheng2020panoptic,chen2020semi,yu2020bisenet,fu2019adaptive}, we use the same online hard example mining (OHEM) strategy during training.
We also find applying the exponential moving average decay~\cite{tan2019efficientnet,tan2020efficientdet} to be helpful to stabilize training without influencing the final performance.

We report our main results in mIoU (Mean Intersection-Over-Union) and pixAcc (Pixel Accuracy) under the single-model single-scale setting with no test-time augmentation. We note that multi-scale inference may further boost the metrics, but they are also much slower in practice.

\paragraph{CityScapes:} CityScapes dataset~\cite{Cordts2016Cityscapes} contains 5,000 high-resolution images with fine annotations, which focuses on urban street scenes understanding. The fine-annotated images are divided into 2,975/500/1,525 images for training, validation and testing respectively. 19 out of 30 total classes are used for evaluation. Following the training protocol from \cite{zhao2017pyramid,sun2019high}, we randomly crop the image from 1024$\times$2048 to 512$\times$1024 for training. We also adopt an initial learning rate of 0.08 and a weight decay of 0.00005. Our models are trained for 90k steps (484 epochs), which is the same as popular methods \cite{sun2019high} (484 epochs).

\paragraph{ADE20K:} ADE20K dataset~\cite{zhou2017scene} is a scene parsing benchmark with 20,210 images for training and 2,000 images for validation.  Models will be evaluated using the mean of pixel-wise accuracy and mIoU over 150 semantic categories. Following previous work~\cite{yuan2019object,zhang2020resnest}, we use the same image size 520$\times$520 for both training and inference. We train our model for 300 epochs with an initial learning rate of 0.08 and a weight decay of 0.00005.

\subsection{Large-Scale Segmentation Results}

We scale up our {\name} to larger network sizes using the scaling configurations in Table~\ref{tab:scaling}. Table~\ref{tab:main} and Table~\ref{tab:ade} compare their performance with other state-of-the-art segmentation models on CityScapes and ADE20K. In general, our {\name} consistently outperforms previous models, often by a large margin. For example, {\name}-S achieves higher accuracy than the latest Auto-DeepLab-S~\cite{liu2019auto} while using 1.5x fewer parameters and 9.7x fewer FLOPs. On CityScapes, {\name}-L achieves 82.6\% mIOU with single-scale, outperforming prior art of HRNetV2-W48 by 1.5\% mIOU with much less FLOPs. These results suggest that {\name} is scalable and can achieve remarkable performance across various resource constraints.
On ADE20K, {\name}-L also outperforms previous multi-scale state-of-the-art CNN models by 1.3\%, while recent transformer-based models achieved better performance with much more computation.

\paragraph{CityScapes test set performance.} Previous works usually retrain or finetune their models on the train+validation set of CityScapes dataset to boost their test set performance. However, such methods introduce extra training cost and more tunable hyperparameters, which is not scalable and leads to harder comparison. Meanwhile, some leaderboard submissions introduce extra tricks such as multi-scale inference and extra post-processing. Here we believe the best practice is to directly evaluate the trained checkpoints on the CityScapes test set, under the simplest single-scale inference protocol. Our {\name}-L shows great generalization ability by achieving a remarkable 84.1\% mIoU on CityScapes leaderboard without bells and whistles, which is on par with top leaderboard results.

\begin{table}
\begin{center}
\resizebox{1.0\columnwidth}{!}{
\begin{tabular}{l|cc}
\hline
Model & mIoU & PixAcc \\
\hline
\textbf{{\name}-M} & \textbf{46.0} & \textbf{81.3} \\
OCR (ResNet-101)~\cite{yuan2019object} & 44.3/45.3$\dagger$ & - \\
HRNetV2-W48~\cite{sun2019high} & 43.1/44.2$\dagger$ & - \\
Auto-DeepLab-M~\cite{liu2019auto} & 42.2$\dagger$ & 81.1$\dagger$ \\
PSPNet (ResNet-101)~\cite{zhao2017pyramid} & 42.0$\dagger$ & 80.6$\dagger$ \\
Auto-DeepLab-S~\cite{liu2019auto} & 40.7$\dagger$ & 80.6$\dagger$ \\
\hline
\textbf{{\name}-L} & \textbf{48.2} & 81.8 \\
DeepLabV3 (ResNeSt-101)~\cite{zhang2020resnest} & 46.9$\dagger$ & 82.1$\dagger$ \\
ACNet (ResNet-101)~\cite{fu2019adaptive} & 45.9$\dagger$ & 82.0$\dagger$ \\
OCR (HRNet-W48)~\cite{yuan2019object} & 44.5/45.5$\dagger$ & - \\
OCNet (ResNet-101)~\cite{yuan2018ocnet} & 45.5$\dagger$ & - \\
DeepLabV3 (ResNeSt-50)~\cite{zhang2020resnest} & 45.1$\dagger$ & 81.2$\dagger$ \\
Auto-DeepLab-L~\cite{liu2019auto} & 44.0$\dagger$ & 81.7$\dagger$ \\
\hline
\textcolor{gray}{SETR~\cite{zheng2021rethinking}} & \textcolor{gray}{46.3} & \textcolor{gray}{-} \\
\textcolor{gray}{Swin-S~\cite{liu2021swin}} & \textcolor{gray}{49.3$\dagger$} & \textcolor{gray}{-} \\
\textcolor{gray}{SegFormer-B4~\cite{xie2021segformer}} & \textcolor{gray}{50.3} & \textcolor{gray}{-} \\
\end{tabular}
}
\end{center}
\vskip -0.2in
\caption{\textbf{Performance comparison on ADE20K.} $\dagger$ denotes results using multi-scale evaluation protocol. All our models are evaluated in single-scale evaluation protocol. Recent Transformer-based models are marked in gray.}
\label{tab:ade}
\vspace{-1em}
\end{table}

\subsection{Real-Time Segmentation Results}
We evaluate our {\name} under real-time inference settings. Since our real-time models are extremely small, we use larger crop size for training, following common practice~\cite{hong2021deep}. Figure~\ref{tab:realtime} shows the performance comparison between our {\name}-Lite and other real-time segmentation models. We measure our inference speed on a Tesla V100-SXM2 16GB GPU with TensorRT-7.2.1.6, CUDA 11.0 and CUDNN 8.0. For fair comparison with the strongest prior art FasterSeg~\cite{chen2019fasterseg}, we use its official open-sourced code, and rerun the inference benchmark under our environment. We use a batch size of 1 and input resolution of 1024$\times$2048. As shown in this table, our {\name}-Lite models outperform the SOTA by a large margin. In particular, our {\name}-Lite-S achieves the highest 189 FPS speed with 2.9\% better mIoU than FasterSeg.  Moreover, our {\name}-Lite-L achieves 80.1\% mIoU on CityScapes, largely bridging the accuracy gap between real-time models and previous full-size models (see Table~\ref{tab:main} for comparison).

\begin{table}[h]
\begin{center}
\begin{tabular}{l|c|c|c}
\hline
Model & mIoU & InputSize & FPS \\
\hline
\textbf{{\name}-Lite-M} & \textbf{74.0} & 512$\times$1024 & \textbf{334}$\dagger$ \\
\hline
\textbf{{\name}-Lite-S} & \textbf{76.0} & 1024$\times$2048 & \textbf{189}$\dagger$ \\
FasterSeg~\cite{chen2019fasterseg} & 73.1 & 1024$\times$2048 & 170$\dagger$ / 164 \\
DF1-Seg~\cite{li2019partial} & 74.1 & 768$\times$1536 & 106 \\
BiSeNet-V2~\cite{yu2020bisenet} & 73.4 & 512$\times$1024 & 156 \\
DF1-Seg-d8~\cite{li2019partial} & 72.4 & 768$\times$1536 & 137 \\
\hline
\textbf{{\name}-Lite-M} & \textbf{78.6} & 1024$\times$2048 & \textbf{125}$\dagger$ \\
DDRNet-23-Slim~\cite{hong2021deep} & 77.8 & 1024$\times$2048 & 109 \\
DF1-Seg2~\cite{li2019partial} & 76.9 & 768$\times$1536 & 56 \\
FANet-34~\cite{hu2020real} & 76.3 & 1024$\times$2048 & 58 \\
DF1-Seg1~\cite{li2019partial} & 75.9 & 768$\times$1536 & 67 \\
BiSeNetV2-L~\cite{yu2020bisenet} & 75.8 & 512$\times$1024 & 47 \\
SwiftNetRN-18\cite{orsic2019defense} & 75.4 & 1024$\times$2048 & 40 \\
FANet-18~\cite{hu2020real} & 75.0 & 1024$\times$2048 & 72 \\
\hline
\textbf{{\name}-Lite-L} & \textbf{80.1} & 1024$\times$2048 & \textbf{79}$\dagger$ \\
DDRNet-23~\cite{hong2021deep} & 79.5 & 1024$\times$2048 & 39 \\
\hline
\end{tabular}
\end{center}
\vskip -0.1in
\caption{\textbf{Performance comparison on CityScapes under real-time speed settings.} $\dagger$ denotes our measurements on the same V100 GPU with the same inference code.}
\label{tab:realtime}
\end{table}

\begin{figure}[t]
\begin{center}
	\includegraphics[width=0.95\columnwidth]{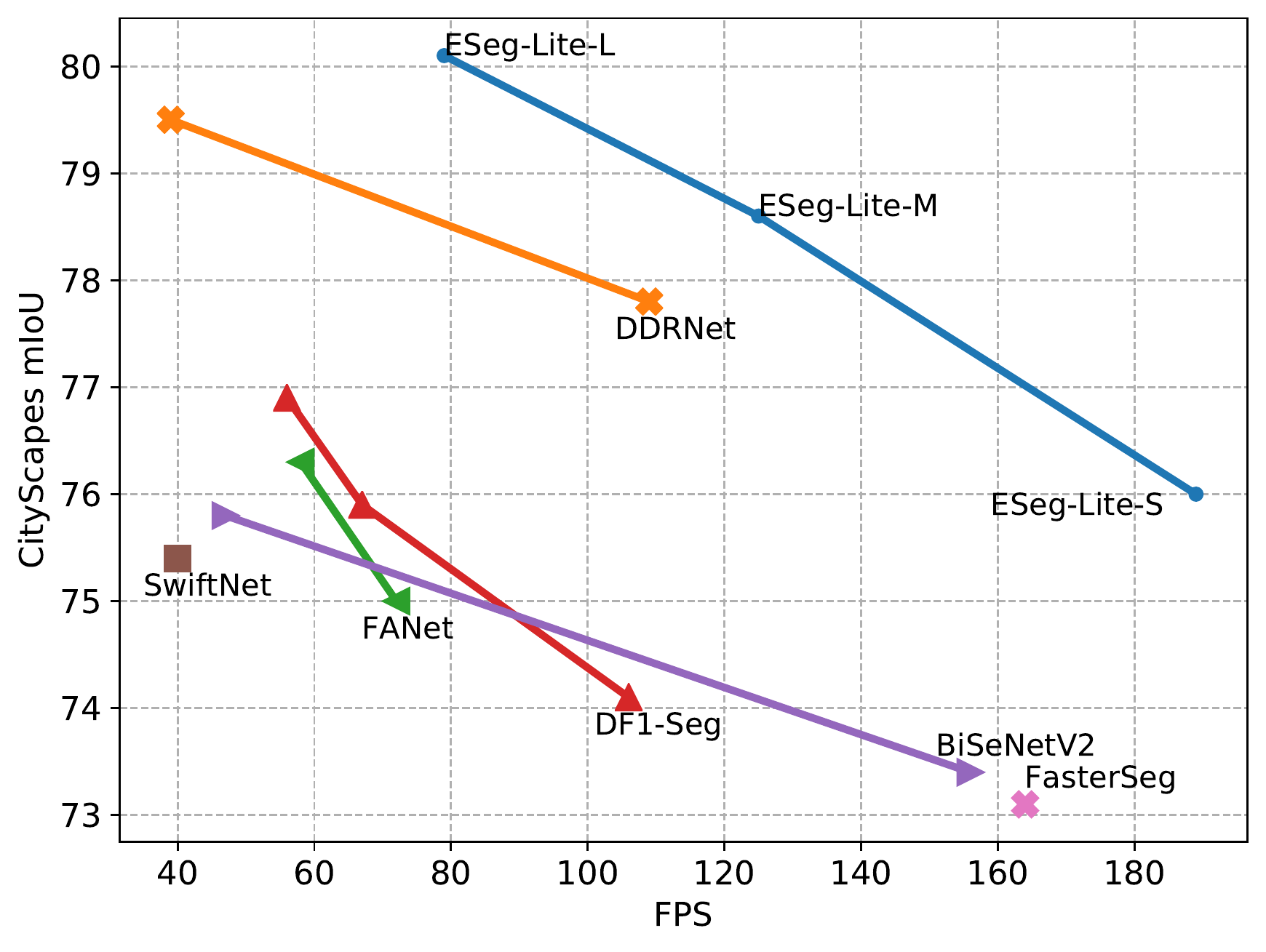}
\end{center}
\vskip -0.1in
	\caption{\textbf{Inference speed vs. CityScapes validation mIoU}. Real-time {\name} family of models outperform previous models by a large margin with much faster speed.}
\vspace{-1em}
\label{fig:teaser}
\end{figure}

\subsection{Pre- and Self-Training with Extra Data}

Given that segmentation datasets often have limited training sets, 
we study two types of data size scaling: pretraining with the large-scale Mapillary Vistas dataset and self-training with CityScapes unlabelled data. Due to the dataset policy, we don't conduct data size scaling experiments on ADE20K dataset~\cite{zhou2017scene}.

\paragraph{Pretraining.} We follow the common setting in \cite{cheng2020panoptic,yuan2019object} to pretrain our models on the larger urban scene understanding dataset Mapillary Vistas~\cite{neuhold2017mapillary}. We resize the images to 2048 pixels at the longer side to handle the image size variations, and randomly crop the images to 1024$\times$1024 during training. The models are pretrained for 500 epochs on Mapillary Vistas dataset. We take a slightly different training setting for finetuning the models from Mapillary Vistas pretrained checkpoints. Specially, we train the models for only 200 epochs with an initial learning rate of 0.01.

\paragraph{Self-training.} We perform self-training on the coarse-annotated data of CityScapes. Note we treat this set as unlabeled data and do not utilize its coarse-annotated labels. To generate the pseudo labels from a teacher, we employ a multi-scale inference setting to generate our pseudo-labels with scales [0.5, 1.0, 2.0] and horizontal flip. We only keep the predictions which have a confidence higher than 0.5, otherwise we set it to ignore area. We use a similar self-training setting as in \cite{zoph2020rethinking}. We use batch size 128 and initial learning rate 0.16 to train the model for 1000 epochs. Half of the images in a batch will be sampled from ground-truth fine-annotated labels, and the other half will be sampled from generated pseudo-labels. Our data scaling method consists of both pretraining and self-training.

\paragraph{} Table~\ref{tab:scaleresults} shows the pretraining and self-training results. Not surprisingly, both pretraining and self-training can improve the accuracy. In general, self-training generally provides slightly higher gains than pretraining for most models, and the accuracy gain on the large model tend to be more pronounced, suggesting that larger models have more network capacity to benefit from the extra data. Interestingly, our results show that large models can obtain more relative accuracy gains with pretraining and self-training. Perhaps surprisingly, we observe that pretraining and self-training are complementary, and combining them can further improve the accuracy by a large margin. For example, {\name}-M can gain 0.9\% accuracy with pretraining and 1.0\% accuracy with self-training, but combining them can provide 2.1\% accuracy gain, significantly better than the single self-training or pretraining. Notably, by combining pretraining and self-training, our largest {\name}-L achieves the best 84.8\% mIoU on CityScapes validation set.

\begin{table}
\begin{center}
\begin{tabular}{l|lll}
\hline
Model size & S & M & L \\
\hline
Baseline & 80.1 & 81.6 & 82.6 \\

Pre-training & 81.1 \footnotesize{(\textcolor{blue}{+1.0})} & 82.5 \footnotesize{(\textcolor{blue}{+0.9})} & 83.8 \footnotesize{(\textcolor{blue}{+1.2})} \\
Self-training & 81.4 \footnotesize{(\textcolor{blue}{+1.3})} & 82.6 \footnotesize{(\textcolor{blue}{+1.0})} & 83.5 \footnotesize{(\textcolor{blue}{+0.9})} \\
Combined & 81.7 \footnotesize{(\textcolor{blue}{\textbf{+1.6}})} & 83.7 \footnotesize{(\textcolor{blue}{\textbf{+2.1}})} & 84.8 \footnotesize{(\textcolor{blue}{\textbf{+2.2}})} \\
\hline
\end{tabular}
\end{center}
\vskip -0.1in
\caption{\textbf{Performance results with pre-training and self-training on various {\name} models.} Observations:   (1)  large  models  like {\name}-XL benefit more from extra data; (2) self-training are  more effective than pretraining; (3) pretraining and self-training are complementary,  and can be combined to obtain the best accuracy gains.}
\label{tab:scaleresults}
\vspace{-1em}
\end{table}

\section{Ablation Study}

In this section, we ablate our encoder and decoder design choices on the CityScapes dataset.

A powerful backbone is crucial for an encoder-decoder architecture as it needs to encode the high-resolution images into low-resolution features. Table \ref{tab:encoder_decoder} shows the performance of different backbones. Compared to the widely-used ResNet-50~\cite{he2016deep}, EfficientNet-B1~\cite{tan2019efficientnet} achieves 1.2\% better mIoU with much fewer FLOPs, suggesting the critical role of good backbone networks in encoder-decoder architectures.

\begin{table}[h]
\begin{center}
\resizebox{1.0\columnwidth}{!}{
\begin{tabular}{l|l|c|c}
\hline
Encoder & Decoder & mIoU & FLOPs \\
\hline
\multirow{3}{*}{EfficientNet-B1} & BiFPN (w/o atrous) & \textbf{80.1} & \textbf{34.5B} \\
	& DeepLabV3+ (w/ atrous) & 79.4 & 91.8B \\
	& DeepLabV3+ (w/o atrous) & 78.8 & 49.9B \\
\hline
\multirow{3}{*}{ResNet-50} & BiFPN (w/o atrous) & 78.9 & 188.0B \\
	& DeepLabV3+ (w/ atrous) & 77.8 & 324.3B \\
	& DeepLabV3+ (w/o atrous) & 77.4 & 230.3B \\
\hline
\end{tabular}
}
\end{center}
\vskip -0.1in
\caption{\textbf{Encoder and decoder choices.} All models are trained with exactly the same training settings. BiFPN  outperforms DeepLabV3+~\cite{chen2018encoder} regardless whether atrous convolutions are used.}
\label{tab:encoder_decoder}
\vspace{-1em}
\end{table}

The decoder also plays an important role as it needs to recover high-resolution features. Table~\ref{tab:encoder_decoder} compares the previously-widely-used DeepLabV3+~\cite{chen2018encoder} with BiFPN (adopted in {\name}). Combined with higher feature levels, the BiFPN achieves better performance than the combination of ASPP and DeepLabV3+ decoder in term of both accuracy and efficiency. Our ablation study also demonstrates the importance of employing high internal resolution in DeepLab-like models, which is very expensive. In contrast, {\name} provides a more elegant and efficient alternative.

%
\section{Related Work}
\paragraph{Efficient Network Architecture:}
Many previous works aim to improve segmentation model efficiency via better hand-crafted networks~\cite{hu2020real,yu2020bisenet} or network pruning~\cite{li2019partial}. Recently, neural architecture search (NAS)~\cite{zoph2016neural,tan2019efficientnet,tan2021efficientnetv2} has become a popular tool to improve model accuracy and efficiency. 
For semantic segmentation, DPC~\cite{chen2018searching} searches for a multi-scale dense prediction cell after a handcrafted backbone; Auto-DeepLab~\cite{liu2019auto} jointly searches for both cell level and network level architecture for segmentation. Recent works~\cite{chen2019fasterseg,zhang2020dcnas} also use a similar hierarchical search space, targeting for real-time speed or better accuracy. %

\paragraph{Encoder-decoder structure:} Early works including U-Net~\cite{ronneberger2015u} and SegNet~\cite{badrinarayanan2017segnet} adopt a symmetric encoder-decoder structure to effectively extract spatial context. Such structure will first extract the high-level semantic information from input images using an encoder network, and then add extra top-down and lateral connections to fuse the features in the decoder part. Most feature encoders used in semantic segmentation come from architectures developed for image classification~\cite{he2016deep,xie2017aggregated,chollet2017xception,zagoruyko2016wide,zhang2020resnest}, while some of them are designed to keep high-resolution features for dense prediction tasks~\cite{li2018detnet,wang2020deep,sun2019high}. Different decoder modules are also proposed to recover high-resolution representations~\cite{ronneberger2015u,noh2015learning,newell2016stacked,yang2017stacked,lin2017feature,pohlen2017full,badrinarayanan2017segnet}. Our work is largely based on encoder-decoder structure for its simplicity.

\paragraph{Multi-scale features: } Multi-scale features have been widely used in dense prediction tasks such as object detection and segmentation. 
While high-resolution features preserve more spatial information with smaller receptive fields,  lower-resolution features can have larger receptive fields to capture more contextual information.
A key challenge here is how to effective combine these multi-scale features. Many previous works in object detection use feature pyramid networks (FPN)\cite{lin2017feature} and its variants~\cite{liu2018path,ghiasi2019fpn} to perform multi-scale feature fusion.
For semantic segmentation, ASPP~\cite{chen2017deeplab} is widely used to get multi-scale features~\cite{zhao2017pyramid,chen2017deeplab,chen2016attention,chen2018encoder}. Other studies also try to exploit the multi-scale features with attention~\cite{yuan2019object,fu2019adaptive}. In our work, we leverage a more powerful bidirectional feature network~\cite{tan2020efficientdet} to extensively fuse multi-scale features.

\paragraph{Atrous convolutions:}
Atrous convolutions is widely used in semantic segmentation architectures~\cite{yu2017dilated,chen2014semantic,chen2017deeplab,zhao2017pyramid,chen2017rethinking,chen2018encoder} to enlarge the receptive field. PSPNet~\cite{zhao2017pyramid} uses a pyramid pooling module on top of a backbone with dilation. DeepLab~\cite{chen2017deeplab} proposes an atrous spatial pyramid pooling to extract multi-scale features for dense prediction. DeepLabV3+~\cite{chen2018encoder} further improves it by using an additional decoder module to refine the representation. 
Unfortunately, atrous convolutions are not hardware friendly and usually run slowly in real-world scenario.
As opposed to the common practice, our proposed model does not use any atrous convolutions.
 
\section{Conclusion}
In this paper, we present a family of simple {\name} models, which use neither high internal resolution nor atrous convolutions. Instead, they add a much richer multi-scale feature space and use a more powerful bi-directional feature network to capture the local and global semantic information. We show that the simple encoder-decoder-based {\name} can outperform the prior art on various datasets. In particular, our small models {\name}-Lite significantly bridge the gap between real-time and server-size models, and our {\name} achieves state-of-the-art performance on both CityScapes and ADE20K with much fewer parameters and FLOPs. With extra pretraining and self-training, our {\name}-L further pushes the CityScapes accuracy to 84.8\% while being single-model and single-scale. Our study shows that, despite the difficulty of dense prediction, it is possible to achieve both high accuracy and faster speed. We hope our work can spark more explorations on how to design segmentation models with better performance and efficiency.

{\small
\bibliographystyle{ieee_fullname}
\bibliography{egbib}
}

\end{document}